\documentclass[letterpaper,comsoc]{IEEEtran}

\usepackage[T1]{fontenc}

\usepackage{cite}
\usepackage{amsmath,amssymb,amsfonts}
\usepackage{algorithmic}
\usepackage{graphicx}
\usepackage{textcomp}
\usepackage{xcolor}

\usepackage{tabularx} 
\usepackage{multirow} 
\usepackage{rotating} 
\usepackage{xcolor,colortbl} 
\usepackage[linesnumbered,ruled,vlined]{algorithm2e}

\SetCommentSty{mycommfont}
\SetKwInput{KwInput}{Input}                
\SetKwInput{KwOutput}{Output}              
\SetKwInput{Init}{Initialization}

\ifCLASSINFOpdf
\else
\fi
\usepackage{amsmath}
\usepackage[cmintegrals]{newtxmath}
\begin{document}
%

%
\title{A Novel Resource Allocation for Anti-jamming in Cognitive-UAVs: an Active Inference Approach}

%
%
%

\author{Ali~Krayani,~\IEEEmembership{Graduate Student Member,~IEEE,}
        Atm~S.~Alam,~\IEEEmembership{Member,~IEEE,}
        Lucio~Marcenaro,~\IEEEmembership{Senior Member,~IEEE,}
        Arumugam~Nallanathan,~\IEEEmembership{Fellow,~IEEE,}
        and~Carlo~Regazzoni,~\IEEEmembership{Senior Member,~IEEE}
        %
%
%
%


%
%


\thanks{Ali Krayani is with DITEN, University of Genoa, 16145 Genoa, Italy, and also with EECS, Queen Mary University of London, London E1 4NS, U.K. (e-mail: ali.krayani@edu.unige.it, a.krayani@qmul.ac.uk). Lucio Marcenaro and Carlo Regazzoni are with DITEN, University of Genoa, 16145 Genoa, Italy, and the Italian National Consortium for Telecommunications (CNIT). (e-mails: 
\{lucio.marcenaro, carlo.regazzoni\}@unige.it). Atm~S.~Alam and Arumugam Nallanathan are with EECS, Queen Mary University of London, London E1 4NS, U.K. (e-mails: \{a.alam, a.nallanathan\}@qmul.ac.uk).
}
}

%
%

\markboth{IEEE Communications Letters}%
{Shell \MakeLowercase{\textit{et al.}}: Bare Demo of IEEEtran.cls for IEEE Communications Society Journals}
%



\maketitle

\begin{abstract}
This work proposes a novel resource allocation strategy for anti-jamming in Cognitive Radio using Active Inference (\textit{AIn}), and a cognitive-UAV is employed as a case study.
An Active Generalized Dynamic Bayesian Network (Active-GDBN) is proposed to represent the external environment that jointly encodes the physical signal dynamics and the dynamic interaction between UAV and jammer in the spectrum. We cast the action and planning as a Bayesian inference problem that can be solved by avoiding surprising states (minimizing abnormality) during online learning.
Simulation results verify the effectiveness of the proposed \textit{AIn} approach in minimizing abnormalities (maximizing rewards) and has a high 
convergence speed by comparing it with the conventional Frequency Hopping and Q-learning.
\end{abstract}
\vspace{-0.1cm}

\vspace{-0.3cm}

\section{Introduction}
With the integration of Unmanned Aerial Vehicles (UAVs), Wireless Communications (WCs) are more prone to terrestrial jammers due to the high heterogeneity and dominant Line-of-Sight (LoS) links \cite{8883124}. Jammers cause damage to communication and degrade the system's performance. Therefore, it is crucial to develop an anti-jamming strategy to reach robust connectivity and improve communication security. 

Cognitive Radio is a key technology to accomplish intelligent resource management in jamming scenarios.
In detecting the existence of the jammers and avoiding jamming attacks, conventional anti-jamming solutions that use fixed transmission patterns can be used. However, they are unable to deal with dynamic jamming patterns in complicated radio environments with high uncertainty, and unpredictable jamming behaviours \cite{9237733}. Recently, Reinforcement Learning (RL) has attracted much attention in WCs to design anti-jamming solutions in complex environments. RL methods such as Q-learning (QL) \cite{7636793} are used to deal with different types of jammers. However, they suffer from slow convergence if the state and action spaces are large, which leads to anti-jamming performance degradation. Deep-QL has been proposed in \cite{7952524} to overcome that issue and learn efficient defence policy. RL methods are based on a reward signal coming from the environment as a feedback to evaluate the performed action. However, defining a proper reward function in complex and dynamic environments is a big challenge \cite{Deep_RL_reward}. \textbf{Active Inference (\textit{AIn})} \cite{6767058} can overcome this challenging task by replacing reward functions with prior beliefs about desired sensory signals received from the environment. Thus, \textit{AIn} agent can learn to describe how it expects itself to behave without getting a feedback from the environment. $\textit{AIn}$ is a promising emerging theory from cognitive neuroscience; it provides a theoretical Bayesian framework that supports how biological agents perceive and act in the real world through the free-energy principle and offers an alternative to RL. 

This letter proposes an \textit{AIn} framework as a novel resource allocation strategy for anti-jamming and studies the Cognitive-UAV based scenario. Under the \textit{AIn} framework, the Cognitive-UAV is endowed with a joint internal representation (generative model) of the external environment, encoding the physical signal and the available physical resources jointly. This enables encoding the dynamic interaction between the UAV and the jammer in the spectrum. The objective is to learn the best set of actions performed by the UAV as interaction with a jammer that leads to the minimum surprise (positive reward). Such a representation goes over the necessity of mapping actions to signals' states directly (unlike the RL approach) and modelling them over a continuous state-space, which can be a complicated task in RL. There are four main rationals to use \textit{AIn} approach over RL (\hspace{-0.009mm}\cite{7636793,7952524}): \textit{i}) \textit{AIn} operates in a pure belief-based setting allowing one to seek information about the environment and resolve uncertainty in a Bayesian-optimal fashion. \textit{ii}) \textit{AIn} enables speeding up the learning process by performing multiple updates simultaneously while adapting to the dynamic changes in the spectrum. \textit{iii}) There is a dynamic balance between the exploration and exploitation due to the pure belief-based mode, while RL is driven by a value function that updates a single state action at each step. \textit{iv}) In \textit{AIn} the reliance on an explicit reward signal coming from the environment is not necessary; the reward is substituted by Generalized Errors that can be treated as self-information to avoid surprising states (i.e., states under attack) and reach the equilibrium. \textit{To our best knowledge, this is the first work that adopts AIn for anti-jamming in WCs.}

\section{System Model and Problem Formulation} 
Consider a cellular-connected UAV communicating with its respective Ground Base Station (GBS) to receive the tele-commands during a given mission of duration $\mathrm{T}$ over the Command and Control (C2) link which does not exceed a data rate of $100$ Kbps \cite{9311617}, while a malicious terrestrial jammer transmits jamming signals with the intention of disturbing the legitimate UAV communications. The jammer may adopt constant, random or sweep jamming patterns during a certain experience. The UAV, GBS and jammer are denoted as $u$, $g$ and $j$, respectively. The 3D coordinate of GBS and jammer are fixed at $\boldsymbol{o^{g}}=[x^{g}, y^{g}, z^{g}]$ and $\boldsymbol{p^{j}}=[x^{j}, y^{j}, z^{j}]$, respectively, while the time-varying coordinate of UAV at time instant $t$ is defined as $\boldsymbol{q^{u}_{t}}=[x^{u}_{t}, y^{u}_{t}, z^{u}_{t}]$.
The path-loss model from the ground equipment (i.e., GBS or jammer) to UAV follows the cellular to UAV path-loss model, which can be expressed according to \cite{8048502} as: $\mathrm{PL^{e,u}_{t}}(d_{t}, \theta_{t}) = \mathrm{PL^{ter}}(d_{t})+\eta(\theta_{t})+\chi(\theta_{t}),$ where $e \in \{g,j\}$, $\mathrm{PL_{t}^{ter}}(d_{t})=10\alpha\log(d_{t})$ is the terrestrial path-loss of the point beneath the UAV, $\alpha$ is the terrestrial path-loss exponent that depends on the propagation environment and $d_{t}=\sqrt{(x^{u}_{t}-x^{e})^{2}+(y^{u}_{t}-y^{e})^{2}}$ is the 2D distance between $e$ and $u$. In addition, $\eta(\theta_{t})=C(\theta_{t}-\theta_{0})\exp{\big(-\frac{\theta_{t}-\theta_{0}}{D}\big)}+\eta_{0}$ is the excess aerial path-loss and $\chi(\theta_{t})$ is a zero-mean Gaussian variable with an angle-dependent standard deviation describing the shadowing effect such that $\chi(\theta_{t}) \sim \mathcal{N}(0, \sigma(\theta_{t}){=}a\theta_{t}+\sigma_{0})$, where $C$ is the excess path-loss scaler, $D$ is the angle scaler, $\theta_{0}$ is the angle offset, $\eta_{0}$ is the excess path-loss offset, $a$ is the UAV shadowing slope, $\theta_{t}=\arctan{\big(\frac{z^{u}_{t}-z^{e}_{t}}{d_{t}}\big)}$ is the depression angle and $\sigma_{0}$ is the UAV shadowing offset.
The GBS assigns one Physical Resource Block (PRB) to the UAV each $t$ where C2 data are transmitted \cite{krayani_globecom}. The set of available links is denoted as $\mathcal{RB}$=$\{f_{1}, \dots, f_{n}, \dots, f_{N}\}$, $1$ $\leq$ n $\leq$ $N$, where $|\mathcal{RB}|$=$N$ is the total number of available PRBs that depends on the channel bandwidth ${BW}$. To cope with the malicious jamming, the UAV aims to learn the best allocation strategy online by selecting the proper PRBs that are not targeted by the jammer while interacting with the environment and sending updated information to GBS to adapt to the environmental dynamic changes. Denote $\mathcal{H}_{0}$ and $\mathcal{H}_{1}$ as the hypotheses of the absence (i.e., UAV and jammer selected different PRBs) and presence (i.e., UAV and jammer selected the same PRB) of the jammer, respectively. The complex signal that is received at the UAV at time instant $t$ and over $f_{n}$ is given as $r_{t,f_{n}}=h_{t,f_{n}}^{g,u}x_{t,f_{n}}^{u} + v_{t}$ and $r_{t,f_{n}}=h_{t,f_{n}}^{g,u}x_{t,f_{n}}^{u}+h_{t,f_{n}}^{j,u}x_{t,f_{n}}^{j}+ v_{t}$ at hypotheses $\mathcal{H}_{0}$ and $\mathcal{H}_{1}$, respectively, where $x_{t,f_{n}}^{u}$ denotes the C2 signal, $h_{t,f_{n}}^{g,u}={1}/{\mathrm{PL_{t}}^{g,u}}$ is the channel gain from GBS to UAV, $x_{t,f_{n}}^{j}$ stands for the jammer's signal, $h_{t,f_{n}}^{j,u}={1}/{\mathrm{PL_{t}}^{j,u}}$ is the channel gain from jammer to UAV and $ v_{t}$ is the random noise. The corresponding SINR at the UAV is given by $\gamma_{t} = {P^{u}_{t}h_{t,f_{n}}^{g,u}}/{(\alpha P^{j}_{t}h_{t,f_{n}}^{j,u}+\sigma^{2})}$, where $P^{u}_{t}$ is the transmitted power, $P^{j}_{t}$ is the jammer power, whose presence is denoted by $\alpha$ which is equal to $0$ under $\mathcal{H}_{0}$ and equals to $1$ under $\mathcal{H}_{1}$.

The anti-jamming defense problem can be formulated as a partially observable Markov decision process (POMDP) since the spectrum is only partially observable to the UAV.
A discrete-time POMDP that models the relationship between the UAV and its environment can be described as 7-element tuple ($\boldsymbol{S}, \boldsymbol{X}, \boldsymbol{\mathcal{A}}, \boldsymbol{\mathcal{P}_{\tau}^{u}}, \boldsymbol{\mathcal{P}_{\tau}^{j}}, \boldsymbol{\Pi_{\tau}^{a^{u}}}, \boldsymbol{{\tilde{Z}_{t,f_{n}}}}$), where $\boldsymbol{S}$ and $\boldsymbol{X}$ are sets of the environmental hidden states, $\boldsymbol{\mathcal{A}}$ is a set of actions where action is PRB selection ($a_{t} \in \mathcal{RB}$), $\boldsymbol{\mathcal{P}_{\tau}^{u}}$ and $\boldsymbol{\mathcal{P}_{\tau}^{j}}$ are the time-varying transition models for UAV and jammer, respectively. $\boldsymbol{\Pi_{\tau}^{a^{u}}}$ is the \textit{AIn}-table that encodes the state-action couple and $\boldsymbol{{\tilde{Z}_{t,f_{n}}}}$ are the observations received at each $t$ over $f_{n}$. During the offline training, UAV learns a dynamic model $\mathcal{M}$ encoding the dynamic rules that generate desired sensory signals (i.e., without jamming interference). During the active inference process (i.e., online learning), UAV predicts the environmental hidden states characterized by the posterior distributions $\mathrm{P}(s^{*}_{t}{\in}\boldsymbol{S}|z_{t}{\in}\boldsymbol{{\tilde{Z}_{t,f_{n}}}}, \mathcal{M}$) and $\mathrm{P}(x^{*}_{t}\in \boldsymbol{X}|z_{t}\in\boldsymbol{{\tilde{Z}_{t,f_{n}}}}, \mathcal{M})$ based on a prior belief (encoded in $\mathcal{M}$) and infers the actions most likely to generate preferred sensory signals (i.e., clean signals without jamming interference). Then, UAV can evaluate the situation after receiving the current observation $z_{t}$ and calculate the similarity between predictions and observations using a probabilistic distance $\mathcal{D}$ (i.e., abnormality indicator). If the similarity is high (i.e., $\mathcal{H}_{0}$), UAV can understand that the selected action has led to desired states and to the reception of desired signals.
If the similarity is low (i.e., $\mathcal{H}_{1}$), UAV can understand that the selected action is a bad action and updates $\boldsymbol{\Pi_{\tau}^{a^{u}}}$ accordingly to avoid selecting actions that lead to surprising states (i.e., high abnormality). Therefore, while acting and sensing the spectrum, the UAV aims to minimise the cumulative abnormality:
\begin{equation}\label{prob_form_minAbnormality_discrete}
\scriptsize
    \min_{a_{t}} \sum_{t=1}^{\mathrm{T}} \mathcal{D_{}}\bigg(\mathrm{P}(s^{*}_{t}|z_{t}, \mathcal{M}), \mathrm{P}(z_{t}|s^{*}_{t}, \mathcal{M})\bigg).
\end{equation}
It is to note that \eqref{prob_form_minAbnormality_discrete} is equivalent to maximize the SINR.
\section{Proposed Anti-jamming method}
\subsection{Radio Environment Representation}
We assume that the environment is described by a Generalized-state-space model, comprised of:
\begin{equation}\label{discrete_model}
\scriptsize
    \tilde{S}_{t,f_{n}}^{u}=\mathrm{F}(\tilde{S}_{t-1,f_{n}}^{u})+\tilde{w}_{t,f_{n}},
\end{equation}
\begin{equation}\label{continuous_model}
\scriptsize
    \tilde{X}_{t,f_{n}}^{u}=A\tilde{X}_{t-1,f_{n}}^{u}+BU_{\tilde{S}_{t,f_{n}}^{u}}+\tilde{w}_{t,f_{n}},
\end{equation}
\begin{equation} \label{observation_model}
\scriptsize
    \tilde{Z}_{t,f_{n}}=H\tilde{X}_{t,f_{n}}^{u}+H\tilde{X}_{t,f_{n}}^{j}+\tilde{\upsilon}_{t,f_{n}},
\end{equation}
In \eqref{discrete_model}, $\tilde{S}_{t,f_{n}}^{u}$ are discrete random variables (or Generalized superstates GSS) describing the discrete clusters of the UAVs' C2 signals that evolve according to \eqref{discrete_model} where $\mathrm{F}(.)$ is a non-linear function describing the signals' dynamic transitions among the discrete variables and its evolution over time at a specific PRB ($f_{n}$) and $\tilde{w}_{t,f_{n}}$ is a Generalized process noise such that, $\tilde{w}_{t,f_{n}} {\sim} \mathcal{N}(0, \Sigma_{\tilde{w}_{t,f_{n}}})$.
The dynamic model in \eqref{continuous_model} explains the dynamic evolution of the continuous random variables $\tilde{X}_{t,f_{n}}$ (or Generalized states GS) where $A {\in} \mathbb{R}^{2d,2d}$, $B {\in} \mathbb{R}^{2d,2d}$ are the dynamic model and control model matrices, respectively, and $U_{\tilde{S}_{t,f_{n}}^{u}}$ is the control vector.
The observation model is given in \eqref{observation_model} where $\tilde{Z}_{t,f_{n}} {\in} \mathbb{R}^{2d}$ is the generalized observations including the signals' features in terms of $I$ and $Q$ components and the $1^{st}$-order temporal derivatives ($\dot{I}$, $\dot{Q}$) where $d$ is the space dimensionality. We assume that each sensory signal is a linear combination of one hidden GS ($\tilde{X}_{t,f_{n}}^{u}$) affected by additive random noise in a normal situation (i.e., under $\mathcal{H}_{0}$) and by additional interference ($\tilde{X}_{t,f_{n}}^{j}$) caused by the jammer in an abnormal situation (i.e., under $\mathcal{H}_{1}$). $\tilde{X}_{t,f_{n}}^{u}$ and $\tilde{X}_{t,f_{n}}^{j}$ are the UAV's GS and the jammer's GS (that is caused by $\tilde{S}_{t,f_{n}}^{j}$), respectively. $H {\in} \mathbb{R}^{2d,2d}$ maps hidden states to observations, $f_n$ is the $n$-th PRB where $f_n {\in} \mathcal{RB}$ and $\tilde{\upsilon}_{t,f_{n}}{\sim}\mathcal{N}(0, \Sigma_{\tilde{\upsilon}_{t,f_{n}}})$. 

\subsection{Offline learning of desired observations}
\begin{figure}[t!]
    \centering
     \includegraphics[height=5.4cm]{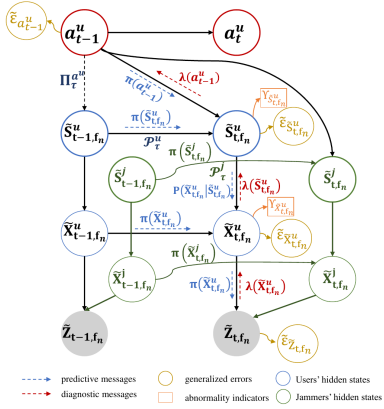}
    \caption{Graphical representation of the proposed Active-GDBN. \color{black} The top-level of the hierarchy stands for the active states ($a_{t-1}^{u}$) representing the actions that the UAV can perform. The UAV can predict the consequences of the performed actions that affect the hidden environmental states ($\mathrm{\tilde{{S}}_{t,f_{n}}}$, $\mathrm{\tilde{{X}}_{t,f_{n}}}$) causing sensory signals ($\mathrm{\tilde{Z}}_{t,f_{n}}$). $\mathrm{\tilde{{S}}_{t,f_{n}}}$ are discrete variables representing the clusters and $\mathrm{\tilde{{X}}_{t,f_{n}}}$ are continuous variables representing the dynamics of the physical signal inside a certain cluster. Edges represent the conditional dependencies among random variables at multiple levels. Each level of the hierarchy holds beliefs about the variables of the level below. Beliefs are signalled via predictive messages in a top-down manner and compared against sensory signals, resulting in multi-level abnormality indicators and generalized errors that are fed back via diagnostic messages in a bottom-up manner. \color{black}}
    \label{fig_GDBN_initial_and_ref}
    \vspace{-0.3cm}
\end{figure}
During training, we assume that the jammer is absent and the UAV aims to learn the dynamics of the desired observations (i.e., C2 signals without jamming interference) while sensing the spectrum.
UAV starts perceiving the surroundings by partially sensing the spectrum, supposing that no signals are present and observations are subject to a stationary noise process that evolves according to static rules. UAV relays on $\eqref{continuous_model}$ to predict the continuous signal's state where the force at sensing PRB ($f_{n}$) is $U_{\tilde{S}_{t,f_{n}}}$=$0$, as no rules have been discovered yet. In case of active transmissions in $f_{n}$, UAV detects abnormalities all the time and calculates the Generalized Errors (GEs) projected on the GS space as follows:
%
%
%
$\bf{\tilde{\mathcal{E}}_{\tilde{X}_{t,f_{n}}^{u}}} {=}$$ \big[\tilde{X}_{t,f_{n}}^{u}, \mathrm{P}(\dot{\mathcal{E}}_{\tilde{X}_{t,f_{n}}^{u}}) \big]{=}
    \big[\tilde{X}_{t,f_{n}}^{u}, H^{-1}\tilde{\mathcal{E}}_{\tilde{Z}_{t,f_{n}}} \big],$
where $\dot{\mathcal{E}}_{\tilde{X}_{t,f_{n}}^{u}}$ is the difference between predictions and observations that capture the dynamics of the signals present inside the spectrum and should be applied to $\tilde{X}_{t,f_{n}}^{u}$ and $\tilde{\mathcal{E}}_{\tilde{Z}_{t,f_{n}}}$=$\tilde{Z}_{t,f_{n}}-H\tilde{X}_{t,f_{n}}^{u}$. 
GEs can be clustered in an unsupervised manner using the Growing Neural Gas (GNG) to learn the top level of abstraction (semantic level).
GNG produces a set of GSS (or clusters) encoding the GEs into discrete regions described by the set $\boldsymbol{\tilde{S}_{f_{n}}^{u}}$, such that:
$\boldsymbol{\tilde{S}_{f_{n}}^{u}}$=$\{\tilde{S}_{1,f_{n}}^{u}, \tilde{S}_{2,f_{n}}^{u}, \dots, \tilde{S}_{M,f_{n}}^{u}\},$
where $M$ is the total number of clusters associated with a specific PRB.
Analysing the signal's dynamic transitions among the GSS and how they vary with time allows estimating the time-varying transition probabilities $\pi_{if_{n}|jf_{n},\tau}^{u}$=$\mathrm{P}(\tilde{S}_{t,f_{n}}^{u}$=$i|\tilde{S}_{t-1,f_{n}}^{u}$=$j, \tau)$ which is encoded in the time-varying transition matrix $\Pi_{f_{n},\tau}^{u}$ where $i,j {\in} \boldsymbol{\tilde{S}_{f_{n}}^{u}}$.
Moreover, each discrete variable $\tilde{S}_{m,f_{n}}^{u} {\in} \boldsymbol{\tilde{S}}_{f_{n}}^{u}$ is associated with statistical proprieties as generalized mean $\tilde{\mu}_{\tilde{S}_{m,f_{n}}^{u}}$ and covariance $\Sigma_{\tilde{S}_{m,f_{n}}^{u}}$.
During offline learning, UAV has been trained to learn and encode the dynamic rules that generate desired sensory signals (i.e., without jamming attacks) using multiple observations (over multiple RBs).
\subsection{Active Inference stage (online learning)}
The hierarchical dynamic models formulated in terms of stochastic processes as defined in \eqref{discrete_model},\eqref{continuous_model},\eqref{observation_model} are structured in an Active Generalized Dynamic Bayesian Networks (Active-GDBN) depicted in Fig.1. The Active-GDBN allows to solve the POMDP to find the best set of actions by predicting the situation the UAV could encounter in the future, conditioned on the actions it executes.
Thus, \textit{AIn} provides a way, through planning as inference, to form beliefs about the future and describe the causal relationship among actions, hidden states and outcomes at multiple levels.
\subsubsection{\textbf{Initialization}}
$\boldsymbol{\mathcal{P}_{\tau}^{u}}$ and $\boldsymbol{\mathcal{P}_{\tau}^{j}}$ are the $N {\times} N$ time-varying matrices encoding the possible transitions among the $N$ available resources performed by the UAV and encoding the UAV's belief about the possible actions that the jammer can perform, respectively. 
Since there is no a priori information concerning the jammer's behaviour inside the spectrum, the probability entries in both $\boldsymbol{\mathcal{P}_{\tau}^{u}}$ and $\boldsymbol{\mathcal{P}_{\tau}^{j}}$ are initially assigned equal values:
\begin{equation}
\tiny
\begin{split}
    \boldsymbol{\mathcal{P}_{\tau}^{u}} = 
    \begin{bmatrix} 
    \mathrm{P}(\Pi_{f_{1}|f_{1},\tau}^{u}) & \dots & \mathrm{P}(\Pi_{f_{1}|f_{N},\tau}^{u}) \\
    \vdots & \ddots & \vdots \\
    \mathrm{P}(\Pi_{f_{N}|f_{1},\tau}^{u}) & \dots & \mathrm{P}(\Pi_{f_{N}|f_{N},\tau}^{u}) 
    \end{bmatrix},
    \boldsymbol{\mathcal{P}_{\tau}^{j}} = 
    \begin{bmatrix} 
    \mathrm{P}(\Pi_{f_{1}|f_{1},\tau}^{j}) & \dots & \mathrm{P}(\Pi_{f_{1}|f_{N},\tau}^{j}) \\
    \vdots & \ddots & \vdots \\
    \mathrm{P}(\Pi_{f_{N}|f_{1},\tau}^{j}) & \dots & \mathrm{P}(\Pi_{f_{N}|f_{N},\tau}^{j}) 
    \end{bmatrix},
    \end{split}
\end{equation}
where $\mathrm{P}(\Pi_{f_{r}|f_{q},\tau}^{u})$=$\frac{1}{N}$, $\mathrm{P}(\Pi_{f_{r}|f_{q},\tau}^{j})$=$\frac{1}{N}$ $\forall r,q {\in} \mathcal{RB}$.
$\boldsymbol{\Pi_{\tau}^{a^{u}}} {\in} \mathbb{R}^{N,N}$ is a time-varying matrix encoding the probabilistic dependencies between states and actions representing the link $a^{u}_{t-1} {\rightarrow} \tilde{S}_{t-1,f_{n}}^{u}$ in the Active-GDBN that describes $\mathrm{P}(a^{u}_{t-1}$=$f_{i}|\tilde{S}_{t-1,f_{k}}^{u})$ and defined as:
\begin{equation}
\tiny
 \begin{split}
    \boldsymbol{\Pi_{\tau}^{a^{u}}} = 
    \begin{bmatrix} 
    \mathrm{P}(a_{1}=f_{1}|\tilde{S}_{t-1,f_{1}}^{u}) & \dots & \mathrm{P}(a_{N}=f_{N}|\tilde{S}_{t-1,f_{1}}^{u}) \\
    \vdots & \ddots & \vdots \\
    \mathrm{P}(a_{1}=f_{1}|\tilde{S}_{t-1,f_{N}}^{u}) & \dots & \mathrm{P}(a_{N}=f_{N}|\tilde{S}_{t-1,f_{N}}^{u})
    \end{bmatrix},
 \end{split}
\end{equation}
where $\mathrm{P}(a^{u}_{t-1}$=$f_{i}|\tilde{S}_{t-1,f_{k}}^{u})$=$\frac{1}{N} \ \forall i,k {\in} \mathcal{RB}.$
UAV's action depends on the state-action couple encoded in $\Pi_{\tau}^{a^{u}}$ and on its belief about the presence of the jammer in the radio spectrum encoded in $\mathcal{P}_{\tau}^{j}$.

\subsubsection{\textbf{Action selection process}}
Initially, UAV performs random sampling to select the actions during the $1^{st}$ iteration as every possible action has the same probability ($\frac{1}{N}$) of being chosen. The selected action $a^{u}_{t-1}$ indicates what will be the next hidden state $\tilde{S}_{t,f_{n}}^{u}$ according to $\mathrm{P}(\tilde{S}_{t, f_{n}}^{u}|\tilde{S}_{t-1,f_{n}}^{u}, a^{u}_{t-1})$. $\tilde{S}_{t,f_{n}}^{u}$ encodes the predicted cluster of the model and the activated PRB ($f_{n}$). 

In the successive iterations, first, UAV predicts the future activity of the jammer implicitly according to $\mathcal{P}_{\tau}^{u}$. Then, it can adjust the action selection step by skipping the risky resources (i.e., resources expected with high probability to be targeted by the jammer in the near future). 
The action selection procedure depends on a certain policy adopted by the UAV according to:
\begin{equation} \label{select_action}
    a_{t-1}^{u*} = \operatorname{argmax}\limits_{\tilde{S}_{t-1,f_{k}}^{u}, \mathcal{P}^{u}_{\tau}(\tilde{S}_{t-1,f_{k}}^{u})} \pi(a_{t-1}^{u}),
\end{equation}
where $\pi(a_{t-1}^{u})$=$\mathrm{P}(a^{u}_{t-1}|\tilde{S}_{t-1,f_{k}}^{u})$ is a specific row in ${\Pi_{\tau}^{a^{u}}}$ and $\mathcal{P}^{u}_{\tau}(\tilde{S}_{t-1,f_{k}}^{u})$ is a specific row selected from ($\mathcal{P}^{u}_{\tau}$) representing the dynamic model associated with ($\tilde{S}_{t-1,f_{k}}^{u}$) where the jammer's transitions are implicitly encoded.
The model has prior belief about how a certain state ($\tilde{S}_{t-1,f_{k}}^{u}$) will evolve into another ($\tilde{S}_{t,f_{k}}^{u*}$) depending on the chosen action ($a_{t-1}^{u*}$) according to:
%
%
$\mathrm{P}(\tilde{S}_{t,f_{k}}^{u*}|a_{t-1}^{u*}, \tilde{S}_{t-1,f_{k}}^{u})$,
where $\tilde{S}_{t,f_{k}}^{u*}$ is the expected state associated with the selected action.

\subsubsection{\textbf{Perception and joint state-prediction}}
After selecting the action that indicates the chosen PRB, UAV can rely on the corresponding transition matrix ($\Pi_{f_{r}|f_{q},\tau}^{u}$) to perform the predictions by employing the Modified Markov Jump Particle Filter (M-MJPF) \cite{krayani_globecom}, that uses a combination of Particle Filter (PF) and a bank of Kalman Filters (KFs).
PF starts by propagating $L$ particles equally weighted based on the proposal density encoded in $\Pi_{f_{r}|f_{q},\tau}^{u}$, such that:
$<$$\tilde{S}^{u,l}_{t,f_{n}}, W^{l}_{t}$$>$${\sim}$$<$$\pi_{if_{n}|jf_{n}, \tau}^{u}, \frac{1}{L}$$>$.
For each particle $\tilde{S}^{u,l}_{t,f_{n}}$, a KF is employed to predict $\tilde{X}_{t,f_{n}}^{u}$. The prediction at this level is driven by the higher level as pointed out in \eqref{continuous_model} (where $U_{\tilde{S}_{t,f_{n}}^{u}}$=$\tilde{\mu}_{\tilde{S}^{u,l}_{t,f_{n}}}$) which can be expressed as $\mathrm{P}(\tilde{X}_{t,f_{n}}^{u}|\tilde{X}_{t-1,f_{n}}^{u}, \tilde{S}_{t,f_{n}}^{u})$. The posterior probability associated with $\tilde{X}_{t,f_{n}}^{u}$ is given by:
$\pi(\tilde{X}_{t,f_{n}}^{u})$=$\mathrm{P}(\tilde{X}_{t,f_{n}}^{u}, \tilde{S}_{t,f_{n}}^{u}|\tilde{Z}_{t-1,f_{n}})$.

Once a new sensory signal is received, diagnostic messages propagate in bottom-up to adjust the expectations and update belief in hidden variables. Thus, the posterior can be updated using:
$\mathrm{P}(\tilde{X}_{t,f_{n}}^{u}, \tilde{S}_{t,f_{n}}^{u}|\tilde{Z}_{t,f_{n}})$=$\pi(\tilde{X}_{t,f_{n}}^{u}) \lambda(\tilde{X}_{t,f_{n}}^{u}).$
In addition, the likelihood message $\lambda(\tilde{S}_{t,f_{n}}^{u})$ can be used to update the particles' weights according to:
$W^{l}_{t}$=$W^{l}_{t} \lambda(\tilde{S}_{t,f_{n}}^{u}),$
where:
$\lambda(\tilde{S}_{t,f_{n}}^{u}) {=} \lambda(\tilde{X}_{t,f_{n}}^{u}) \mathrm{P}(\tilde{X}_{t,f_{n}}^{u}|\tilde{S}_{t,f_{n}}^{u}) {=} \mathrm{P}(\tilde{Z}_{t,f_{n}}^{u}|\tilde{X}_{t,f_{n}}^{u}) \mathrm{P}(\tilde{X}_{t,f_{n}}^{u}|\tilde{S}_{t,f_{n}}^{u}),$
and $\mathrm{P}(\tilde{X}_{t,f_{n}}^{u}|\tilde{S}_{t,f_{n}}^{u}) {\sim} \mathcal{N}(\mu_{\tilde{S}_{m,f_{n}}^{u}},\,\Sigma_{\tilde{S}_{m,f_{n}}^{u}})$ denotes a multivariate Gaussian distribution.
Also, GE ($\tilde{\mathcal{E}}_{\tilde{S}^{u}_{t,f_{n}}}$) at the superstate level conditioned on transiting from $\tilde{S}^{u}_{t-1,f_{n}}$ can be expressed as:
$\bf{\tilde{\mathcal{E}}_{\tilde{S}^{u}_{t,f_{n}}}}$ ${=} \big[ \tilde{S}_{t-1,f_{k}}^{u}, \dot{\mathcal{E}}_{\tilde{S}^{u}_{t,f_{n}}} \big],$
where $\dot{\mathcal{E}}_{\tilde{S}^{u}_{t,f_{n}}}$ is an aleatory variable whose probability density function is given by $\mathrm{P}(\dot{\mathcal{E}}_{\tilde{S}^{u}_{t,f_{n}}})$=$\lambda(\tilde{S}_{t,f_{n}}^{u}) - \pi(\tilde{S}_{t,f_{n}}^{u})$ representing the new force that can be used to update $\mathcal{P}_{\tau}^{u}$ and thus improve future predictions.

\subsubsection{\textbf{Abnormality measurements}}
In order to evaluate to what extent the current signal's evolution at the discrete level matches the predicted one based on the learned and encoded dynamics in the model, we used an abnormality indicator ($\boldsymbol{\Upsilon_{\tilde{S}_{t,f_{n}}^{u}}}$) based on the Symmetric Kullback-Leibler ($\mathcal{SKL}$) Divergence ($D_{KL}$) \cite{krayani_globecom}. $\boldsymbol{\Upsilon_{\tilde{S}_{t,f_{n}}^{u}}}$ calculates the similarity between the two messages that represent discrete probability distributions entering to node $\tilde{S}_{t,f_{n}}^{u}$, namely, $\pi(\tilde{S}_{t,f_{n}}^{u})$ and $\lambda(\tilde{S}_{t,f_{n}}^{u})$, it is associated with $\bf{\tilde{\mathcal{E}}_{\tilde{S}^{u}_{t,f_{n}}}}$
and formulated as:
\begin{equation}\label{KLDA_abn}
\scriptsize
\begin{split}
    \boldsymbol{\Upsilon_{\tilde{S}_{t,f_{n}}^{u}}} = \sum_{i \in \mathcal{S}} \mathrm{P}_{r}(\tilde{S}_{t,f_{n}}^{u}=i)D_{KL}\big(\pi(\tilde{S}_{t,f_{n}}^{u})||\lambda(\tilde{S}_{t,f_{n}}^{u})\big) +  \\ \sum_{i \in \mathcal{S}} \mathrm{P}_{r}(\tilde{S}_{t,f_{n}}^{u}=i)D_{KL}\big(\lambda(\tilde{S}_{t,f_{n}}^{u})||\pi(\tilde{S}_{t,f_{n}}^{u})\big),
\end{split}
\end{equation}
where $\mathrm{P}_{r}(\tilde{S}_{t,f_{n}}^{u})$ is the probability of occurrence of each superstate picked from the histogram at time instant $t$ and calculated as follows: $\mathrm{P}_{r}(\tilde{S}_{t,f_{n}}^{u})$=$\frac{fr(\tilde{S}_{t,f_{n}}^{u}=i)}{N},$ where $fr(.)$ is the frequency of occurrence of a specific superstate $i$, $N$ is the total number of particles propagated by PF, and $\mathcal{S}$ is the set consisting of all winning particles, such that: $\mathcal{S} = \big\{i|\mathrm{P}_{r}(\tilde{S}_{t,f_{n}}^{u}) > 0 \big\}, \ i \in {\boldsymbol{S^{u}_{f_{n}}}}.$

Likewise, it is possible to understand how much the observation supports the predictions at the GS level using:
\begin{equation}
    \boldsymbol{\Upsilon_{\tilde{X}_{t,f_{n}}^{u}}} = -\ln \bigg( \mathcal{BC}\big(\pi(\tilde{X}_{t,f_{n}}^{u}),\lambda(\tilde{X}_{t,f_{n}}^{u})\big) \bigg),
    \label{CLA_abn_reference_model}
\end{equation}
where $\mathcal{BC}(.)=\int \sqrt{\pi(\tilde{X}_{t,f_{n}}^{u})\lambda(\tilde{X}_{t,f_{n}}^{u})} d\tilde{X}_{t,f_{n}}^{u}$ is the Bhattacharyya coefficient and
$\boldsymbol{\Upsilon_{\tilde{X}_{t,f_{n}}^{u}}}$ is associated with $\bf{\tilde{\mathcal{E}}_{\tilde{X}_{t,f_{n}}^{u}}}$.
\subsubsection{\textbf{Updating of action selection process}}
After acting in the environment, UAV can save the consequence of the chosen action (i.e., the transition from $\tilde{S}_{t-1,f_{k}}^{u}$ to $\tilde{S}_{t,f_{k}}^{u*}$) in $\mathcal{P}^{u}_{\tau}$ and evaluate how much the sensory outcomes support predictions and thus evaluate if the performed action was good or bad by using the abnormality measurements defined in \eqref{KLDA_abn} and \eqref{CLA_abn_reference_model}.
In addition, it is possible to calculate the GE ($\tilde{\mathcal{E}}_{a_{t-1}^{u}}$) during abnormal situations to adapt UAV's strategy in selecting actions and understand how it should behave in the future to avoid the jammer. $\tilde{\mathcal{E}}_{a_{t-1}^{u}}$ is the difference between observation and expectation which can be expressed as:
$\tilde{\mathcal{E}}_{a_{t-1}^{u}} {=} \big[{a_{t-1}^{u*}}, \dot{\mathcal{E}}_{a_{t-1}^{u}} \big]$,
where $\dot{\mathcal{E}}_{a_{t-1}^{u}}$ depicts an aleatory variable representing the new force that should be applied to update $\boldsymbol{\pi}(a_{t-1}^{u})$ and its probability density function is given by $\mathrm{P}(\dot{\mathcal{E}}_{a_{t-1}^{u}})$=$\lambda(a^{u}_{t-1}) - \boldsymbol{\pi}(a_{t-1}^{u})$ that can be used as a metric alternative to the reward in RL.
$\lambda(a_{t-1}^{u})$ is the diagnostic message travelling from $\tilde{S}_{t,f_{n}}^{u}$ towards $a_{t-1}^{u}$ and defined as:
$\lambda(a_{t-1}^{u})$=$\lambda(\tilde{S}_{t,f_{n}}^{u})\mathrm{P}(\tilde{S}_{t,f_{n}}^{u}|a_{t-1}^{u})$ representing a discrete probability distribution that holds information about the observed sensory signal and encoding the probabilities about how the states $\tilde{S}_{t,f_{n}}^{u}$ belonging to the available frequencies change based on the evidence, it is given by:
\begin{equation}
\lambda(a_{t-1}^{u})=
\begin{cases}
    \mathcal{P}_{\tau-1}(\tilde{S}_{t-1,f_{n}}^{u})-\gamma^{*}, \ \textit{if} \ a_{t-1}^{u} = a_{t-1}^{u*},
    \\
    \mathcal{P}_{\tau-1}(\tilde{S}_{t-1,f_{n}}^{u})+\frac{\gamma^{*}}{N-1}, \ \textit{if} \ a_{t-1}^{u} \not= a_{t-1}^{u*},
\end{cases}
\end{equation}
where $\gamma$ depends on the GE ($\bf{\tilde{\mathcal{E}}_{\tilde{S}^{u}_{t,f_{n}}}}$), that is:
$\gamma{=}\gamma^{*}$ \textit{if} $\tilde{\mathcal{E}}_{\tilde{S}_{t,f_{k}}} {\geq} \textit{th},$ and $\gamma {=} 0$ \textit{if} $\tilde{\mathcal{E}}_{\tilde{S}_{t,f_{k}}} {<} \textit{th},$
where $\textit{th}$ is the threshold indicating whether the radio situation is normal or abnormal and the value of $\gamma^{*}$ depends on the abnormality indicators defined in \eqref{KLDA_abn} and \eqref{CLA_abn_reference_model}. Hence, GE ($\tilde{\mathcal{E}}_{a_{t-1}^{u}}$) is proportional to $\bf{\tilde{\mathcal{E}}_{\tilde{S}^{u}_{t,f_{n}}}}$ due to the messages propagated from lower level towards the higher levels, such that $\tilde{\mathcal{E}}_{a_{t-1}^{u}}$=$f(\tilde{\mathcal{E}}_{\tilde{S}^{u}_{t,f_{n}}})$.
When the UAV get surprised by the sensory outcomes after performing a certain action, it can use the prediction error signal to update its belief about the jammer's transition model to improve future actions. The core idea is that the user occupying a piece of the spectrum should minimize the abnormality (surprise) associated with finding itself in unlikely states (states under attack).
Jammer's dynamic model ($\mathcal{P}_{\tau}^{j}$) can be updated following:
\begin{equation}
    \mathcal{P}_{\tau}^{j}(.\ ,\tilde{S}_{t,f_{n}}^{j}) = \mathcal{P}_{\tau-1}^{j}(.\ ,\tilde{S}_{t,f_{n}}^{j}) - \mathrm{P}(\dot{\mathcal{E}}_{a_{t-1}}^{u}),
\end{equation}
In an abnormal situation, the user and jammer share the same RB, which means they performed the same action. Thus, the user should update ${\Pi_{\tau}^{a^{u}}}$ by decreasing the probability of selecting that action as follows:
\begin{equation}
    \pi^{*}(a_{t-1}^{u}) = \pi(a_{t-1}^{u}) + \mathrm{P}(\dot{\mathcal{E}}_{a_{t-1}}^{u}),
\end{equation}
and update $\mathcal{P}_{\tau}^{u}$ by decreasing the probability of transiting to $\tilde{S}_{t,f_{k}}^{u}$ from $\tilde{S}_{t-1,f_{k}}^{u}$ after choosing action $a^{u*}_{t-1}$ using the GE ($\bf{\tilde{\mathcal{E}}_{\tilde{S}^{u}_{t,f_{n}}}}$) following:
\begin{equation}
    \mathcal{P}_{\tau}^{u}(\tilde{S}_{t-1,f_{k}}^{u},\tilde{S}^{u}_{t,f_{n}}) = \mathcal{P}_{\tau-1}^{u}(\tilde{S}_{t-1,f_{k}}^{u},\tilde{S}^{u}_{t,f_{n}}) + \mathrm{P}(\dot{\mathcal{E}}_{\tilde{S}^{u}_{t,f_{n}}}).
\end{equation}

\section{Results and Discussion}
\begin{figure*}[t!]
    \centering
    \begin{minipage}[t]{0.32\linewidth}
    \includegraphics[height=1.75cm]{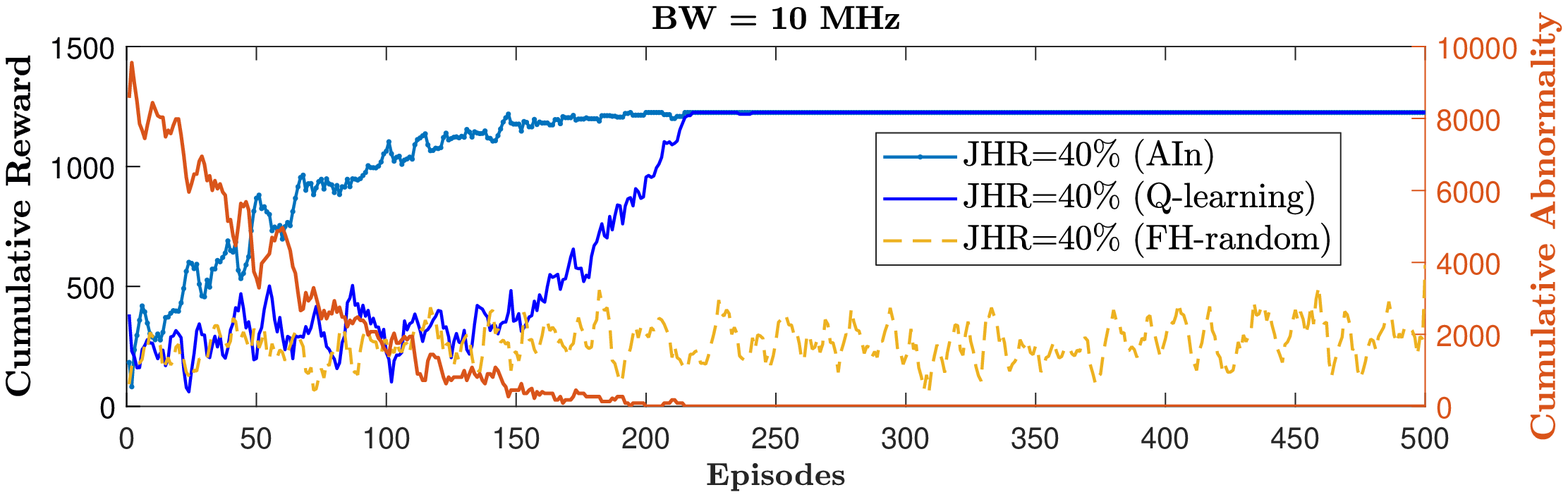}
    \\[-1.9mm]
    \centerline{\scriptsize (a) \textbf{Constant Jammer}}
    \end{minipage}
    \begin{minipage}[t]{0.32\linewidth}
    \includegraphics[height=1.75cm]{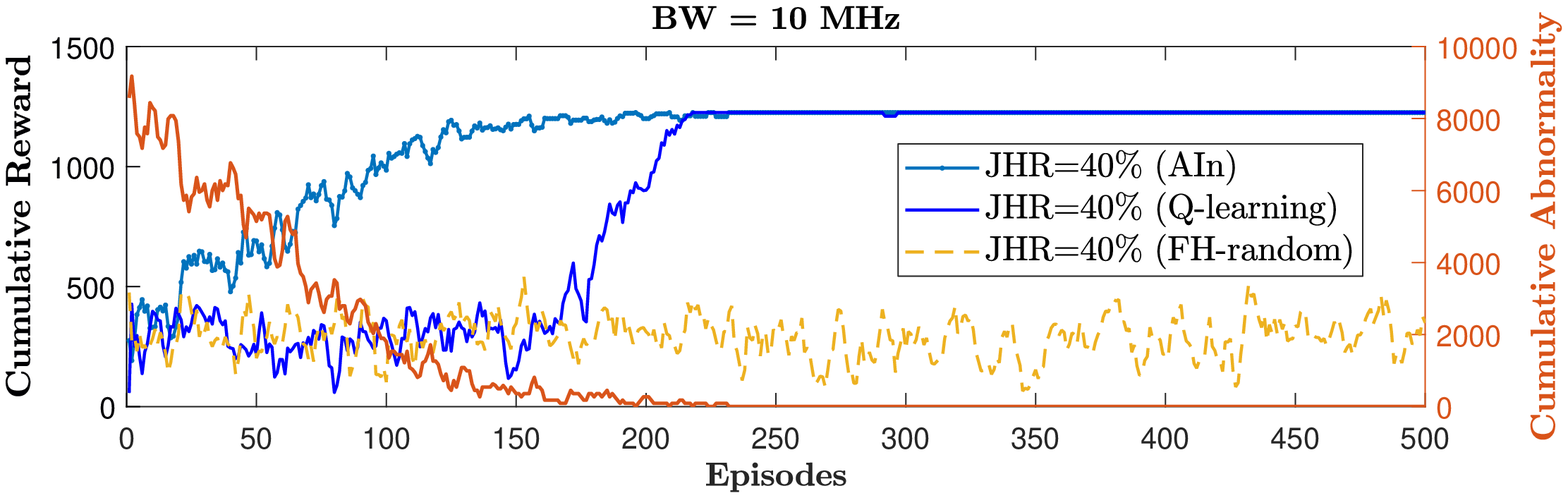}
    \\[-1.9mm]
    \centerline{\scriptsize (b) \textbf{Random Jammer}}
    \end{minipage}
    \begin{minipage}[t]{0.32\linewidth}
    \includegraphics[height=1.75cm]{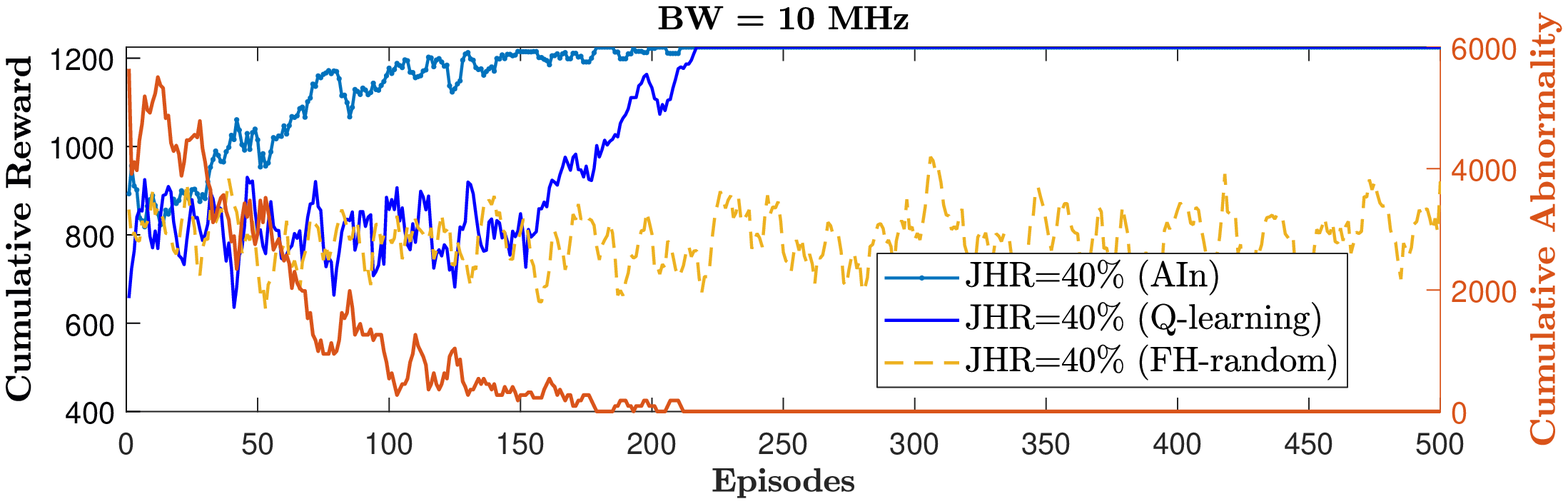}
    \\[-1.9mm]
    \centerline{\scriptsize (c) \textbf{Sweep Jammer}}
    \end{minipage}
    \\[-1.0mm]
    \caption{Performance comparison of cumulative reward and abnormality ($\mathcal{SKL}$) with the proposed \textit{AIn}, FH and QL under different jamming strategies.}
    \label{cum_Abn_Reward_angleChannelModel_allJammers}
    \vspace{-0.1cm}
\end{figure*}
\begin{figure*}[t!]
    \centering
    \begin{minipage}[t]{0.32\linewidth}
    \includegraphics[height=1.75cm]{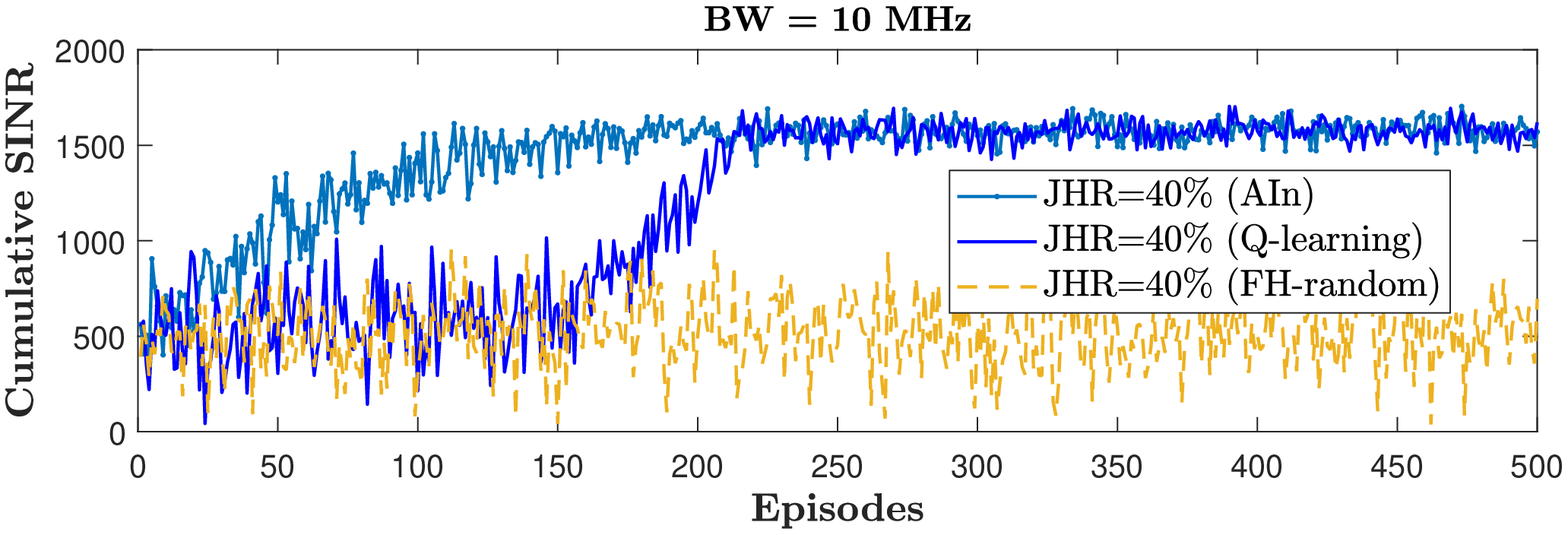}
    \\[-1.9mm]
    \centerline{\scriptsize (a) \textbf{Constant Jammer}}
    \end{minipage}
    \begin{minipage}[t]{0.32\linewidth}
    \includegraphics[height=1.75cm]{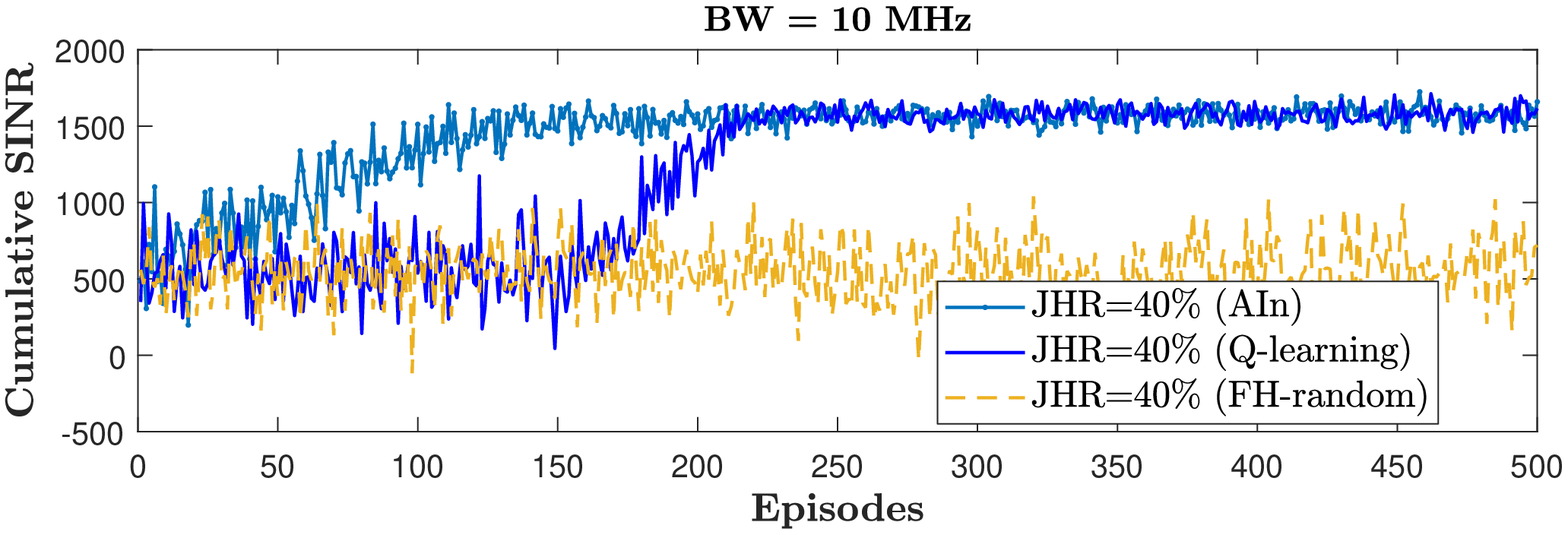}
    \\[-1.9mm]
    \centerline{\scriptsize (b) \textbf{Random Jammer}}
    \end{minipage}
    \begin{minipage}[t]{0.32\linewidth}
    \includegraphics[height=1.75cm]{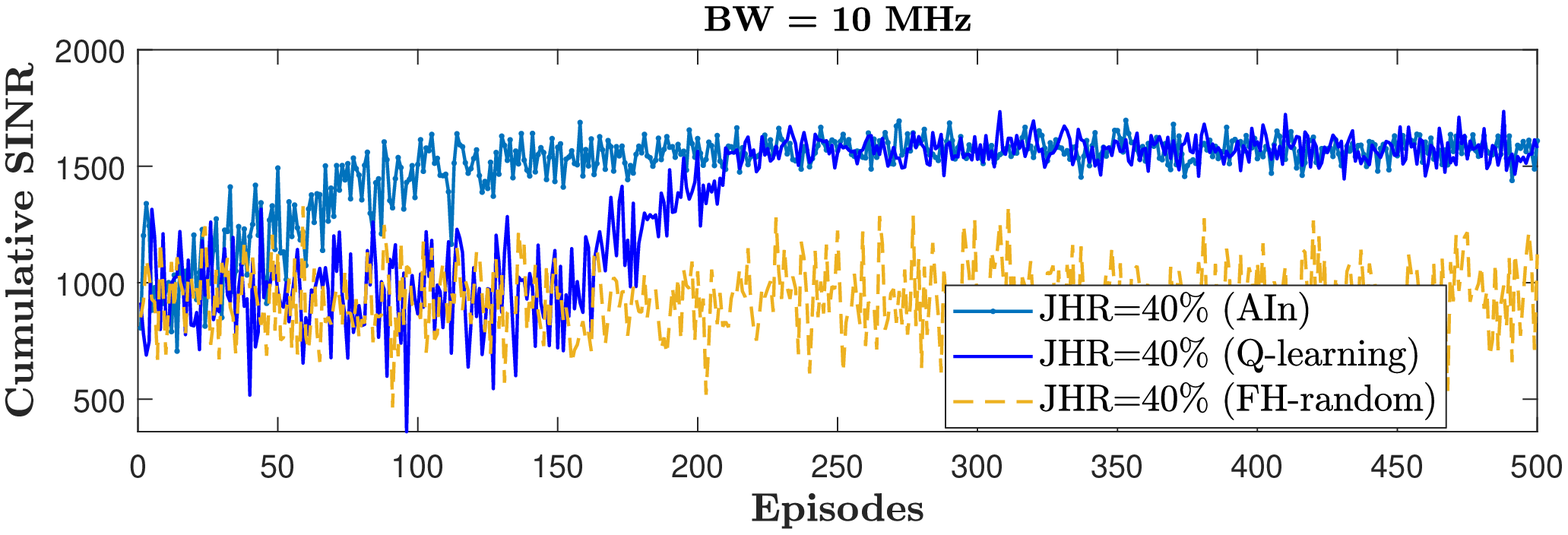}
    \\[-1.9mm]
    \centerline{\scriptsize (c) \textbf{Sweep Jammer}}
    \end{minipage}
    \\[-1.0mm]
    \caption{Performance comparison of cumulative SINR with the proposed \textit{AIn}, FH and QL under different jamming strategies.}
    \label{cum_SINR_angleChannelModel_allJammers}
\end{figure*}
\vspace{-0.2cm}
To evaluate the performance of the proposed \textit{AIn} approach for anti-jamming, following three types of jammers are considered in the simulation:
\textit{1)} Constant jammer that acts on statistically pre-configured channels;
\textit{2)} Sweep jammer that attacks by sweeping among the available PRBs at each time slot; and
\textit{3)} Random jammer that selects uniformly random actions to attack the available PRBs.
The simulation settings are as: BW=$10$MHz; FDD; sub-carrier spacing of $15$ KHz; number of PRBs per BW is $50$; sampling frequency of $1.92$ MHz; $N_{FFT}$ of $128$; $7$ OFDM symbols per slot; normal CP; SNR of $15dB$; QPSK for C2 and jamming signal; jamming to signal power ratio (JSR) of 6dB; and a total of 200 radio frames. In addition, the propagation environment is a typical suburban, mean aerial speed is $4.8$m/s, BS height is $30$m, UAV height is $60$m and the channel model parameters \cite{8048502} are $\alpha{=}3.04$, $\sigma_{0}{=}8.52$, ${C}{=}-23.29$, $\eta_{0}{=}20.70$, $\theta_{0}{=}-3.61$, ${D}{=}4.14$, $a{=}-0.41$, $\sigma_{0}{=}5.86$\color{black}, where a perfect CSI is assumed. \color{black}
Also, we consider a jamming hit rate (JHR) of JHR=$40$\%. C2 data, jamming signals and UAV trajectory are generated as in \cite{krayani_globecom}.

Let us compare the performance of \textit{AIn} in terms of cumulative abnormality (defined in \eqref{KLDA_abn}) and cumulative reward with that of random Frequency Hopping (FH-random) and Q-Learning (QL), as illustrated in Fig.~\ref{cum_Abn_Reward_angleChannelModel_allJammers}. Here, the objective of \textit{AIn} is to minimize abnormality while that of QL is to maximize reward. Thus, the reward is considered in \textit{AIn} approach just for the sake of comparison with QL. We consider a binary reward which is equal to $-1$ under $\mathcal{H}_{1}$ and $+1$ under $\mathcal{H}_{0}$. Nevertheless, the relationship of these metrics is opposites to one another. For a fair comparison with QL, we use time-varying q-tables to deal with the dynamic environmental changes. The exploration process in QL follows the $\epsilon$-greedy policy with $\epsilon=1$ decaying to 0. It can be seen from the figure that \textit{AIn} outperforms QL and FH-random under different jamming strategies while AIn converges faster than QL due to its capability in discovering jammer's policy and performing multiple updates.
Fig.~\ref{cum_SINR_angleChannelModel_allJammers} depicts the cumulative SINR under different jamming patterns achieved by the proposed $\textit{AIn}$ and compared it with FH-random and QL. By observing Fig.~\ref{cum_Abn_Reward_angleChannelModel_allJammers} and Fig.~\ref{cum_SINR_angleChannelModel_allJammers}, we can notice that minimizing the abnormality (or maximizing the reward) leads to maximizing the SINR where the time needed to reach the convergence is equivalent to that in Fig.~\ref{cum_Abn_Reward_angleChannelModel_allJammers} and $\textit{AIn}$ beats both the FH-random and QL. This means that avoiding surprising states minimizes the abnormality and maximises reward and SINR.
\color{black} \textit{AIn} outperforms FH and QL due to its ability to characterize the jammer and discover its attacking strategy, explaining how the UAV should act in the environment. Since \textit{AIn} operates in a pure belief-based setting. It can evaluate whether the action was correct or wrong and also understand how to correct those actions using the errors by performing multiple updates to the \textit{AIn}-table, which speeds up the learning process and reach convergence faster. In contrast, QL performs single updates to the q-table without being able to explain how to correct the wrong actions, hindering the learning process. While FH can not reach convergence as it is always selecting random actions.\color{black}
\vspace{-0.1cm}
\section{Conclusion}
\vspace{-0.2cm}
This letter has proposed a novel resource allocation strategy using Active Inference for anti-jamming in a Cognitive-UAV scenario.
Simulated results have indicated that the proposed method outperforms conventional Frequency Hopping and Q-Learning in terms of learning speed (convergence).
Further research will explore performance improvements by facing smart reactive jammers in fully-observable environments.

\bibliographystyle{IEEEtran}
\bibliography{References}

\vspace{11pt}

\end{document}